\documentclass[conference]{ieeeconf}
\IEEEoverridecommandlockouts

\usepackage{cite}
\usepackage{amsmath,amssymb,amsfonts}
\usepackage{graphicx}
\usepackage{textcomp}
\usepackage{xcolor}
\usepackage{url}
\usepackage{newtxtext,newtxmath}
\usepackage{booktabs}
\usepackage{multirow}
\usepackage{makecell}

\usepackage{caption}
\begin{document}

\title{Deep Reinforcement-Learning-Guided Model Predictive Control for Preventing Overtakes in Autonomous Racing}

\author{
Yufei Xi$^{1\dagger}$, Yijie Liao$^{2\dagger}$, and Tulga Ersal$^{1}$%
\thanks{$^\dagger$These authors contributed equally to this work.}
\thanks{$^{1}$Department of Mechanical Engineering, College of Engineering, University of Michigan, Ann Arbor, MI, USA.}
\thanks{$^{2}$Y. Liao was with the Department of Electrical Engineering and Computer Science, 
College of Engineering, University of Michigan, Ann Arbor, MI, USA. 
He is now with the School of Computing, National University of Singapore, Singapore.}
\thanks{Corresponding author: \texttt{tersal@umich.edu}}
}

\maketitle

\begin{abstract}
This paper addresses defensive blocking in autonomous racing, where a vehicle must prevent a faster opponent from overtaking while operating near its dynamic limits. 
Different from lap-time minimization, we formulate defense as a spatial occupancy regulation problem via a hierarchical reinforcement-learning guided model predictive control framework.
A Soft Actor-Critic strategic layer operates in the Frenet domain to generate geometry-aware defensive references, which are embedded into the nonlinear model predictive control formulation as spatial regularization under friction constraints. 
Evaluated on the Thunderhill West circuit in simulation, the framework increases average overtake time from 8.8 s to 14.6 s while significantly reducing opponent progress.
Meanwhile, it allows the vehicle to utilize 83.4\% of available tire force.
The framework achieves a 33.3 ms mean solve time (13.9 ms std), supporting real-time high-speed adversarial interaction.
\end{abstract}


\section{Introduction}

Autonomous racing has emerged as a demanding benchmark for high-performance control due to high speeds, strong nonlinear tire dynamics, and friction-limited operating conditions \cite{yu_spatial_2025,tubempcracing,subosits_racetrack_2019}. 
Operating near dynamic limits requires precise modeling and rapid decision-making, where small errors can lead to instability. 
When multiple vehicles interact competitively, the problem further extends from single-agent optimal control to multi-agent strategic decision-making under dynamic constraints.

Existing autonomous racing research predominantly focuses on lap-time minimization and overtaking maneuvers. 
Nonlinear model predictive control (MPC) formulations with detailed vehicle models have achieved competitive single-vehicle performance \cite{yu_spatial_2025,YU2021412}. 
Game-theoretic extensions incorporate interaction reasoning to approximate Nash equilibria in multi-vehicle scenarios \cite{wang_game-theoretic_2021,hu2024game}. 
Discrete planners based on Monte Carlo Tree Search have demonstrated interactive behaviors at a high decision level \cite{huang_fair_2025}. 
However, these works primarily aim at maximizing forward progress or overtaking.

In contrast, defensive blocking, where the ego vehicle actively prevents an opponent from overtaking, introduces a fundamentally different objective structure. 
Rather than optimizing pure lap-time performance, the ego vehicle must regulate spatial positioning to restrict the opponent’s feasible overtaking corridor, potentially sacrificing short-term speed to maintain strategic advantage. 
This shifts the optimization focus from self-performance improvement to spatial occupancy control under nonlinear, friction-limited dynamics. 
Despite its relevance in competitive racing and adversarial driving scenarios, defensive behavior remains underrepresented in high-fidelity autonomous racing formulations.

From a methodological perspective, three challenges arise. 
First, at the dynamics level, nonlinear tire-force characteristics and friction constraints demand dynamically feasible control within short horizons \cite{yu_spatial_2025}. 
Second, at the interaction level, many game-theoretic approaches rely on simplified vehicle models or discrete action spaces to maintain computational tractability \cite{wang_game-theoretic_2021,huang_fair_2025}, limiting fidelity in high-speed racing. 
Third, at the algorithmic level, MPCcan ensure constraint satisfaction, but lacks adaptive strategic learning, whereas Reinforcement Learning (RL) captures interaction patterns, but often lacks explicit stability and safety guarantees \cite{wurman_outracing_2022}. 
Pure end-to-end RL approaches have shown competitive performance in simulation \cite{remonda2024simulation}, yet typically do not enforce dynamic constraints explicitly, raising challenges for robustness and sim-to-real transfer in safety-critical racing scenarios. 
Hybrid RL--MPC frameworks \cite{gros_data-driven_2020,rickenbach_zipmpc_2025,amos_differentiable_2019} attempt to combine learning and optimization, but most focus on performance enhancement or model adaptation rather than explicitly modeling defensive spatial interaction under nonlinear dynamics.

To address these gaps, this paper formulates defensive blocking as a spatial strategy learning problem embedded within nonlinear MPC. 
The proposed framework leverages RL to learn geometry-aware defensive intent in the Frenet domain, generating spatial reference points that encode blocking behavior relative to the opponent and track geometry. 
These learned references are transformed into the Cartesian domain and incorporated into a Spatial Envelope MPC formulation as policy-induced regularization terms. 
The MPC layer enforces nonlinear vehicle dynamics and constraint satisfaction, enabling dynamically feasible and real-time defensive control in closed-loop adversarial interaction.

The main contributions of this paper are:
\begin{enumerate}
    \item Formulation of defensive blocking in high-speed, friction-limited autonomous racing as a learned spatial-reference generation problem, addressing competitive interaction beyond lap-time optimization or overtaking.

    \item Embedding of the learned defensive intent into a nonlinear Spatial Envelope MPC through policy-induced spatial regularization, enabling dynamically feasible and real-time closed-loop defensive control.
\end{enumerate}

The remainder of this paper is organized as follows. 
Sec.~\ref{sec:relatedWork} reviews related work in autonomous racing, game-theoretic planning, and hybrid RL--MPC control. 
Sec.~\ref{sec:methodology} presents the proposed framework. 
Sec.~\ref{sec:simulations} details the simulation setup and results. 
Sec.~\ref{sec:conclusions} concludes the paper.

\section{Related Work}
\label{sec:relatedWork}

\subsection{Autonomous Racing and Trajectory Optimization}

MPC is the dominant framework in autonomous racing due to its ability to explicitly handle nonlinear vehicle dynamics and physical constraints. 
Most works focus on lap-time minimization and high-speed trajectory tracking.

Racing is commonly formulated as a nonlinear optimal control problem leveraging detailed tire models and tailored cost functions to maximize acceleration and cornering performance 
\cite{yu_spatial_2025,tubempcracing,YU2021412,neuralnetMPCFriction}. 
Efficient reformulations, including quadratically constrained trajectory replanning, enable real-time execution \cite{subosits_racetrack_2019}. 
Learning has also been incorporated through neural cost maps, online model adaptation, and imitation learning for real-time racing policies \cite{drews_aggressive_2017,costa_online_2023}. 
These methods improve racing performance, but explicit multi-vehicle interaction and defensive positioning are typically outside their scope.

\subsection{Game-Theoretic and Adversarial Planning}

To capture competitive interaction, game-theoretic planning extends optimal control to multi-agent settings. 
Unlike lap-time–oriented formulations, these methods explicitly reason about blocking and overtaking behaviors.

Continuous differential game approaches approximate Nash equilibria within receding-horizon MPC frameworks \cite{wang_game-theoretic_2021}. 
Hierarchical extensions further enable cooperative team strategies through structured interaction reasoning \cite{hu2024game}. 
Such methods embed adversarial reasoning in the control loop, yet often rely on simplified vehicle models and structured track scenarios to maintain tractability.

Discrete planners based on Monte Carlo Tree Search (MCTS) approximate Nash equilibria over discrete action spaces \cite{huang_fair_2025}. 
While effective at high-level strategic reasoning, discretization limits responsiveness under nonlinear high-speed dynamics.

Learning-based interaction modeling has also been explored. 
Neural Nonlinear Opinion Dynamics learned from inverse dynamic games enable rapid overtaking decisions \cite{11127283}, curriculum reinforcement learning has been applied to autonomous overtaking in high-fidelity racing simulation \cite{song2021overtaking,wurman_outracing_2022}, and data-driven priors have been incorporated into dynamic games through Kullback--Leibler regularization \cite{lidard2024blending}. 
These approaches primarily target overtaking, decision-level interaction, or policy-prior shaping rather than embedding defensive spatial intent within nonlinear vehicle models. 
This motivates studying learned defensive spatial references as a way to introduce strategic interaction reasoning into nonlinear MPC.

\subsection{Hybrid RL--MPC Control}

RL enables high-level policy learning through interaction \cite{wurman_outracing_2022}, yet pure RL lacks guarantees on dynamic feasibility and constraint satisfaction. 
Hybrid RL--MPC frameworks address this gap by using RL to approximate economic MPC, compress long-horizon costs, guide hierarchical decisions, or differentiate through the optimizer \cite{gros_data-driven_2020,rickenbach_zipmpc_2025,hori2025learning,amos_differentiable_2019}
Although effective for performance optimization and model adaptation, existing hybrid approaches rarely focus on learning explicit defensive interaction strategies under nonlinear vehicle dynamics.

In contrast, our framework learns \textit{defensive spatial reference points} in the Frenet domain and embeds them into a collocation-based nonlinear MPC. 
Rather than modifying cost parameters or differentiating through the solver, the learned spatial intent acts as policy-induced regularization, bridging strategic defensive reasoning with dynamically feasible control.

\section{Methodology}
\label{sec:methodology}

\subsection{Overall Framework}

\textit{Defensive blocking} is evaluated by overtake time, defined as the first time at which the opponent's Frenet progress exceeds the ego vehicle's by a threshold $\Delta s_{\mathrm{pass}}$ for at least $T_{\mathrm{hold}}$; if no overtake occurs, the value is set to the simulation horizon.

\begin{figure}
    \centering
    \includegraphics[width=\linewidth]{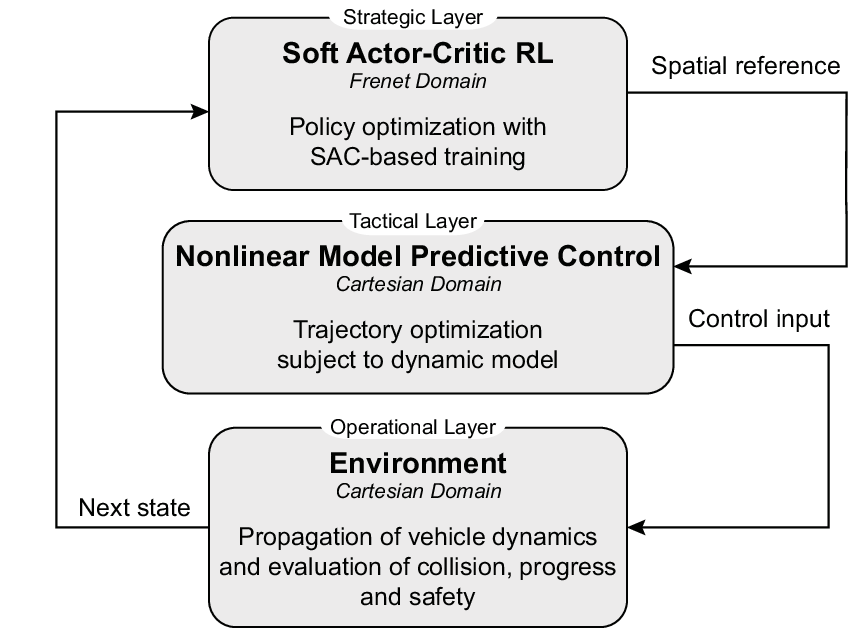}
    \caption{Hierarchical deep RL–MPC framework for adversarial driving.}
    \label{fig:framework}
\end{figure}

This paper proposes the hierarchical RL--MPC framework shown in Fig.~\ref{fig:framework}.
It contains three layers:

\begin{enumerate}
    \item \textbf{Strategic layer:} A deep RL module learns high-level spatial policies in the Frenet coordinate domain, capturing the adversarial interaction and spatial structure of defensive driving. 
    This module is implemented using Soft Actor-Critic (SAC), where both the actor and critic are parameterized by multi-layer neural networks. 
    The actor produces a continuous action distribution that enables expressive nonlinear decision-making, while the critic approximates soft Q-values through deep function approximation. 
    Together, these components provide the capacity needed for modeling complex multi-vehicle interactions and learning robust defensive strategies.

    \item \textbf{Tactical layer:} The learned Frenet-space policy is transformed into the global coordinate frame, establishing the theoretical connection between spatial references and time-parameterized trajectories. 
    A spatial envelope MPC optimizes the control inputs to the vehicle to realize the learned behavior under physical feasibility and safety constraints.

    \item \textbf{Operational layer:} In this layer, the control inputs are applied to a dynamics simulator to obtain the next vehicle state. 
    During this rollout, several metrics are evaluated to quantify task performance. 
    Among them, \emph{collision} indicates whether the ego vehicle makes contact with the opponent, \emph{progress} measures the vehicle's forward advancement along the track, and \emph{safety} assesses whether the vehicle remains within the track boundaries. 
    These evaluated metrics are then passed to the strategic layer, where they are incorporated into the reward function for training.

\end{enumerate}

\subsection{Vehicle Dynamics and Frenet Representation}

In high-speed adversarial racing tasks, the motion of the vehicle is influenced by complex nonlinear factors such as tire forces, coupling between longitudinal and lateral dynamics, and track curvature.

To achieve a balance between physical accuracy and learning generalization, the proposed framework divides vehicle modeling into two complementary parts: a \emph{dynamic model in Cartesian coordinates}, which updates the ego vehicle state by applying the control inputs and simulating its interaction with the opponent for environment and MPC computation, and a \emph{Frenet representation} employed for track-aligned policy learning. 
This separation allows high-level decision-making to generalize across tracks while maintaining physically consistent low-level control.

In the tactical and operational layer, the vehicle dynamics follow the classical nonlinear dynamic single-track model as in \cite{yu_spatial_2025}, with the state and control vectors defined as:
\begin{equation}
    x = \begin{bmatrix} X \\ Y \\ \psi \\ v_x \\ v_y \\ r \\ \delta \\ a_x\end{bmatrix}
    = \begin{bmatrix}
        \text{Global X position} \\
        \text{Global Y position} \\
        \text{Yaw angle} \\
        \text{Longitudinal velocity} \\
        \text{Lateral velocity} \\
        \text{Yaw rate}\\
        \text{Steering angle}\\
        \text{Longitudinal acceleration}
    \end{bmatrix}
    \label{eq1}
\end{equation}

\begin{equation}
    u = \begin{bmatrix} \dot{\delta} \\ j_x \end{bmatrix}
    = \begin{bmatrix}
        \text{Steering rate} \\
        \text{Longitudinal jerk}
    \end{bmatrix}
    \label{eq2}
\end{equation}

Instead of directly using the steering angle $\delta$ and longitudinal acceleration $a_x$ as control inputs, this work adopts their rates as the control input to explicitly capture actuator dynamics and avoid discontinuous control commands, ensuring physically smooth steering and acceleration profiles \cite{yu2024real}.

The continuous-time vehicle dynamics $\dot{x} = f(x,u)$ are then expressed as:
\begin{equation}
    \dot{x} =
    \begin{bmatrix}
        v_x \cos\psi - v_y \sin\psi \\
        v_x \sin\psi + v_y \cos\psi \\
        r \\
        a_x - \frac{1}{m} (F_{yf}\sin\delta + F_{yr}) + v_y r \\
        \frac{1}{m} (F_{yf}\cos\delta + F_{yr}) - v_x r \\
        \frac{1}{I_z} (l_f F_{yf}\cos\delta - l_r F_{yr})\\
        \dot{\delta}\\
        j_x
    \end{bmatrix}
    \label{eq3}
\end{equation}
where $m$ and $I_z$ denote the vehicle mass and yaw moment of inertia, $l_f$ and $l_r$ are the distances from the center of gravity to the front and rear axles, and $\delta$ is the front steering angle. 
$F_{yf}$ and $F_{yr}$ represent the lateral tire forces on the front and rear wheels, whose formulations follow the Fiala tire model \cite{fiala1954seitenkrafte}. 

The model in \eqref{eq3} is employed as the prediction model within the MPC framework. 
The continuous-time dynamics are discretized with a fixed sampling time and imposed as equality constraints in the finite-horizon nonlinear program. 
The rate-based inputs $(\dot{\delta}, j_x)$ are directly optimized, while $\delta$ and $a_x$ are treated as dynamic states evolving through the discretized dynamics, ensuring smooth actuator behavior.

\begin{figure}
    \centering
    \includegraphics[width=0.91\linewidth]{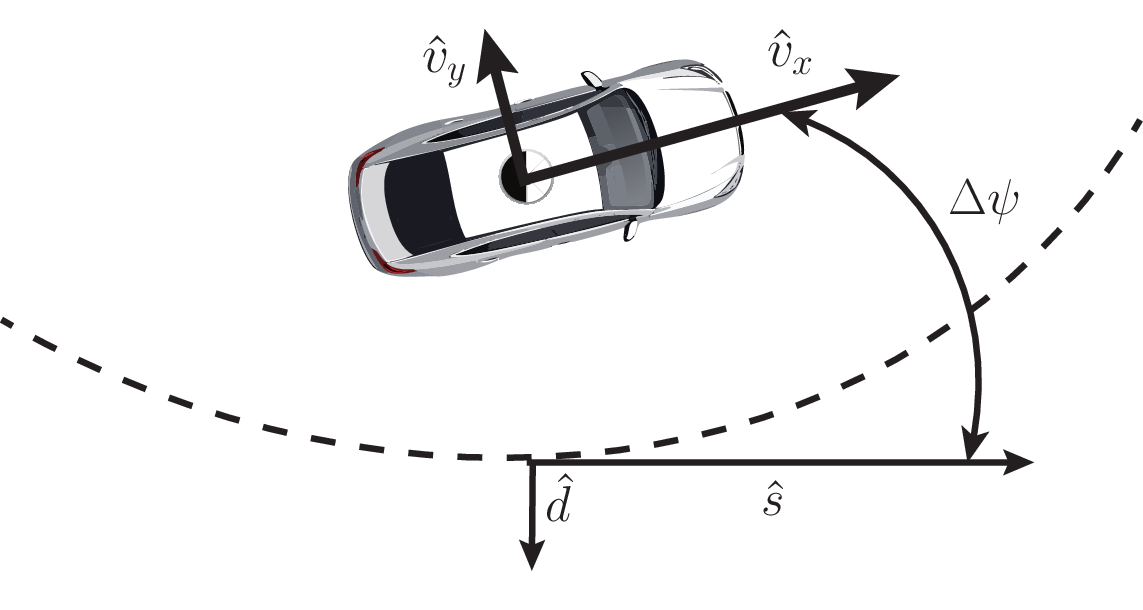}
    \caption{Vehicle representation in the Frenet coordinate frame.}
    \label{fig:frenet}
\end{figure}

However, directly performing policy learning in the global $(X,Y)$ coordinate frame is inefficient and sensitive to absolute position, track geometry, and initialization. 
To address this, the framework adopts a \textit{Frenet frame} in the policy learning layer~\cite{subosits_racetrack_2019}, aligning the vehicle dynamics with the track geometry as shown in Fig.~\ref{fig:frenet}.

The corresponding tangent direction, unit tangent, and unit normal vectors are given by:
\begin{equation}
\begin{cases}
    \psi_c(s) = \arctan\!\left(\frac{dY_c(s)}{dX_c(s)}\right)\\[4pt]
    \mathbf{t}(s) = [\cos\psi_c(s), \sin\psi_c(s)]^\top\\[4pt]
    \mathbf{n}(s) = [-\sin\psi_c(s), \cos\psi_c(s)]^\top
\end{cases}
\label{eq5}
\end{equation}
where $s$ is the arc-length parameter along the centerline, $(X_c(s), Y_c(s))$ denote the global coordinates of the projection point on the track centerline at arc length $s$, and $\psi_c(s)$ is the corresponding heading angle of the centerline.

The global position of the vehicle is therefore expressed as:
\begin{equation}
\begin{bmatrix}
    X \\[3pt] Y
\end{bmatrix}
=
\begin{bmatrix}
    X_c(s) \\[3pt] Y_c(s)
\end{bmatrix}
+ d\,\mathbf{n}(s)
\label{eq6}
\end{equation}
where $d$ is the lateral offset (positive to the right of the track). 

The heading deviation $\Delta\psi$ is defined as
$$\Delta\psi=\psi -\psi_c(s)$$

These geometric tracking variables $(s, d, \Delta\psi)$ are then used to construct track-relative observations for the high-level RL policy. 
The physical system dynamics and the subsequent MPC optimization remain governed by the Cartesian model in \eqref{eq3}. As illustrated in Fig.~\ref{fig:frenet}, the Frenet coordinate system provides a locally aligned spatial reference that facilitates track-consistent representation. 
Using $(s, d, \Delta\psi)$ as the state in the deep RL layer mitigates overdependence on absolute coordinates and ensures invariance across different track layouts and initial positions.

\subsection{Deep Reinforcement Learning Module and Spatial Policy Generation}

In the strategic layer, we employ a SAC framework \cite{haarnoja_soft_2018} to learn a defensive spatial policy in the Frenet frame. Rather than outputting direct control signals, this layer generates a geometrically meaningful spatial reference set, $\mathcal{R}$, which guides the MPC module to ensure physically feasible control.

The spatial reference set is defined as:
\begin{equation}
  \mathcal{R} = \{ (s^{\pi}_i, d^{\pi}_i, w_i) \}_{i=1}^{N_p}
  \label{reference_pos}
\end{equation}
where each point represents an expected lateral offset $d_i$ and an associated weight $w_i$ at a future arc-length $s_i$. The superscript $\pi$ denotes quantities generated by the policy, while $w_i$ is fixed rather than learned.

\subsubsection{State and Action Space}
The state vector $\boldsymbol{\xi}_t$ captures the essential interaction dynamics between the ego ($e$) and opponent ($o$) vehicles in the Frenet frame:
\begin{equation}
\begin{split}
  \boldsymbol{\xi}_t = [\, &s_e, d_e, \Delta\psi_e, v_s^e, v_d^e, \\
  &s_o, d_o, \Delta\psi_o, v_s^o, v_d^o, \Delta s, \Delta d \,]^{\top}
\end{split}
\end{equation}

Given $\boldsymbol{\xi}_t$, the actor network outputs a continuous action vector representing lateral and longitudinal offset corrections at $N_p$ future nodes:
\begin{equation}
  \boldsymbol{\mu}_t = [\Delta s_1,\,\dots,\,\Delta s_{N_p},\, \Delta d_1,\, \dots,\,\Delta d_{N_p}\,]^{\top}
\end{equation}
The spatial reference set is then updated as $\mathcal{R} = \{ (\bar{s}_i+\Delta s_i, d_e+\Delta d_i,w_i)\}_{i=1}^{N_p}$, where $\bar{s}_i$ is the nominal future arc-length node. By operating in the spatial domain, the learned policy achieves high generalization across varying track geometries.

\subsubsection{SAC Optimization}
The agent is trained using standard SAC objectives. The actor network is optimized by maximizing the entropy-regularized expected return ($J_{\pi}$), while the critic network minimizes the Bellman error ($J_Q$) to stabilize value estimation:
\begin{equation}
  J_{\pi} = \mathbb{E}_{\boldsymbol{\xi}_t \sim \mathcal{D}} [\, \alpha \log \pi_{\phi}(\boldsymbol{\mu}_t | \boldsymbol{\xi}_t) - Q_{\theta}(\boldsymbol{\xi}_t, \boldsymbol{\mu}_t) \,]
\end{equation}
\begin{equation}
\begin{split}
  J_Q = \mathbb{E}_{(\boldsymbol{\xi}_t, \boldsymbol{\mu}_t, \boldsymbol{\xi}_{t+1}) \sim \mathcal{D}} \Big[ \big( &Q_{\theta}(\boldsymbol{\xi}_t, \boldsymbol{\mu}_t) \\
  &- ( r_t + \gamma V_{\bar{\theta}}(\boldsymbol{\xi}_{t+1})) \big)^2 \Big]
\end{split}
\end{equation}

\subsubsection{Reward Function}
The reward function $r_t$ balances progress, defensive effectiveness, and safety via fixed weighting coefficients $\lambda_i > 0$:
\begin{equation}
  r_t = \lambda_1 r_{\text{prog}} + \lambda_2 r_{\text{def}} - \lambda_3 r_{\text{coll}} - \lambda_4 r_{\text{bound}}
\end{equation}

\paragraph{Longitudinal progress}
Measured by the forward arc-length increment, accounting for closed-loop track continuity:
\begin{equation}
  r_{\text{prog}} = \mathrm{wrap}(s_t - s_{t-1}).
\end{equation}

\paragraph{Defensive effectiveness}
Encourages the ego vehicle to maintain a forward advantage ($\Delta s_t^{\text{rel}} = \mathrm{wrap}(s_t - s_t^{o})$):
\begin{equation}
  r_{\text{def}} =
  \begin{cases}
  k_1 \Delta s_t^{\text{rel}} + c_1 & \Delta s_t^{\text{rel}} > \epsilon\\
  - c_2 & \text{otherwise}
  \end{cases}
\end{equation}
Sustained loss of this lead triggers an episode termination and an additional penalty.

\paragraph{Collision penalty}
A binary penalty ($r_{\text{coll}} = 1$)  activated if the Cartesian distance between vehicles falls below a safety radius $\delta$, terminating the episode.

\paragraph{Boundary penalty}
A quadratic penalty for exceeding the local track half-width $d_{\max}(s_t)$, normalized by $d_{\text{margin}}$:
\begin{equation}
  r_{\text{bound}} =
  \begin{cases}
  \left(\frac{|d_t| - d_{\max}(s_t)}{d_{\text{margin}}}\right)^2 & |d_t| > d_{\max}(s_t)\\
  0 & \text{otherwise}
  \end{cases}
\end{equation}

\subsection{Deep RL--MPC Integration and Collocation Alignment}
To achieve a differentiable and optimizable coupling between the high-level strategy and the low-level control, 
this work embeds the spatial reference points generated by the deep RL module directly into the nonlinear MPC solving process. 
By aligning selected collocation points with the strategic references in the Frenet domain, 
the controller realizes the learned spatial intent under dynamic feasibility constraints.

The MPC discretizes the prediction horizon into $N$ collocation points in the time domain, 
denoted as $\{t_k\}_{k=0}^{N-1}$. 
During optimization, a subset of collocation nodes $\{N_{k}\}$ is aligned with the deep RL-generated strategic reference points defined in \eqref{reference_pos} by introducing a reference-tracking term in the cost function:
\begin{equation}
J_{\text{ref}} = 
\sum_{i\in {N_{k}}} 
w_i \left(
\| X_{k_i} - X_i^{\pi} \|_{Q_s}^2
+ 
\| Y_{k_i} - Y_i^{\pi} \|_{Q_d}^2
\right)
\label{eq15}
\end{equation}
where $Q_s$ and $Q_d$ are weighting matrices,  $(X_{k_i},Y_{k_i})$ denote the vehicle’s position 
evaluated at each collocation node, and $X_i^{\pi}$, $Y_i^{\pi}$ are the spatial references from deep RL transformed to the global Cartesian coordinates. 
This alignment term effectively attracts the MPC trajectory toward the RL-learned geometric intent while ensuring physical feasibility.

The complete MPC problem is formulated as:
\begin{equation}
\begin{aligned}
\min_{\{x_k, u_k\}} \quad & J = \sum_{k=0}^{N-1} \Big[ \lambda_{\text{ref}} J_{\text{ref}}(x_k) + \lambda_{\text{safe}} J_{\text{safe}}(x_k) \\ & \qquad \qquad + \lambda_{\text{smooth}} J_{\text{smooth}}(x_k, u_k) \Big] \\
\text{s.t.} \quad & x_{k+1} = F(x_k, u_k) \\
& g(x_k) \leq 0 \\
& \left(\frac{F_{x,j}}{\mu_j F_{z,j}}\right)^2+
\left(\frac{F_{y,j}}{\mu_j F_{z,j}}\right)^2 \leq 1,\quad j\in\{f,r\}\\
& x_{\min} \leq x_k \leq x_{\max} \\
& u_{\min} \leq u_k \leq u_{\max}
\end{aligned}
\label{eq16}
\end{equation}
where $J_{\text{safe}}$ denotes the spatial envelope cost encouraging the vehicle 
to remain within track boundaries, and $J_{\text{smooth}}$ represents regularization 
terms on control inputs and state variations. 
The inequality constraints $g(x_k)\leq 0$, friction-circle constraints, and state and input bounds 
$x_{\min}\le x_k \le x_{\max}$ and $u_{\min}\le u_k \le u_{\max}$ 
enforce the physical and track-boundary constraints. 
The weights $\lambda_{\text{ref}}$, $\lambda_{\text{safe}}$, and 
$\lambda_{\text{smooth}}$ balance reference tracking, spatial safety shaping, 
and smoothness regularization, respectively.

Under this structure, the spatial distribution learned by the RL module no longer acts as an external trajectory reference, but becomes a \textit{policy-induced regularization term} embedded within the MPC cost. 
Since the reference alignment is re-evaluated at each MPC iteration, the learned geometric intent continuously influences the optimal solution while respecting dynamic constraints.

\subsection{Benchmark Formulation}
The reference tracking cost, $J_\text{ref}$, serves as the bridge between the high-level RL strategy and the low-level vehicle dynamics. 
To isolate its effect, we define a baseline as Spatial Envelope MPC~\cite{yu_spatial_2025} with $J_\text{ref}$ removed.
This benchmark constitutes an ablation of the proposed framework, and is not intended to represent the strongest possible defensive racing controller.

\section{Simulation Setup and Evaluation}
\label{sec:simulations}
\subsection{Training and Simulation Setup}

\begin{figure}[t]
    \centering
    \includegraphics[width=\linewidth]{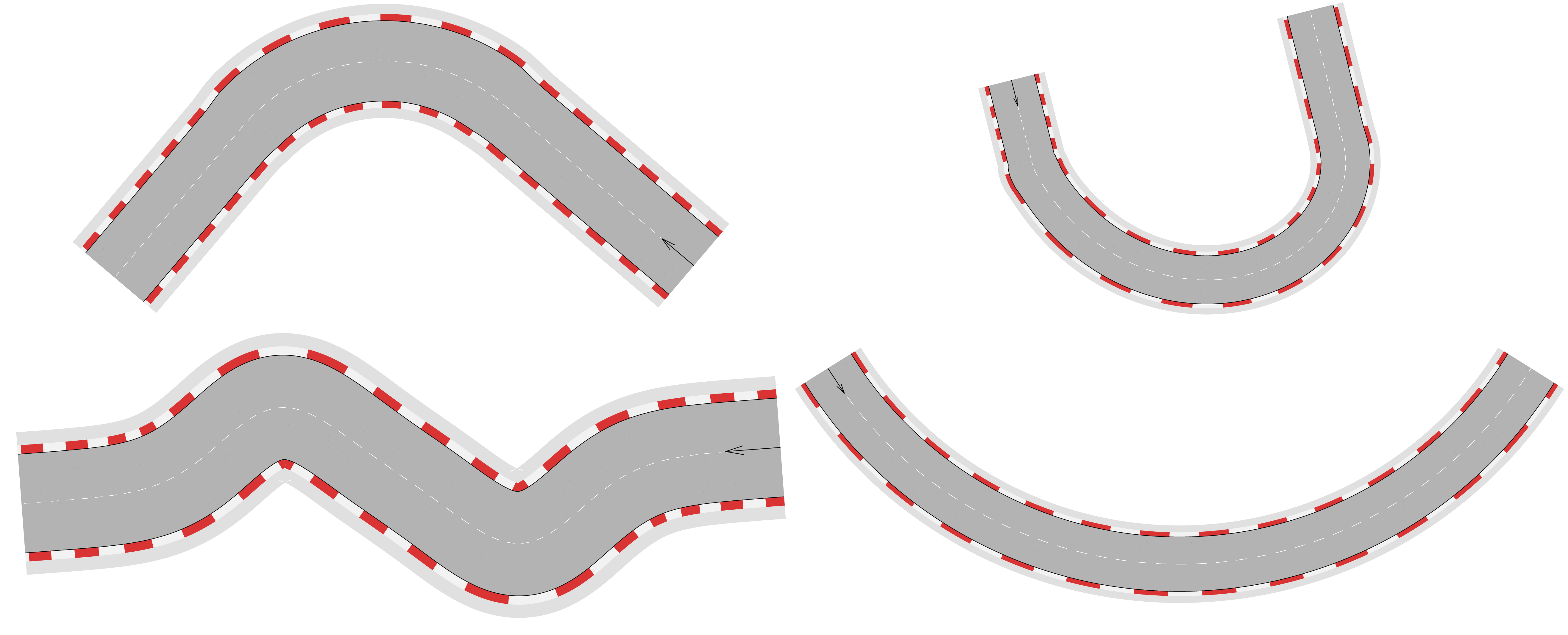}
    \caption{Representative procedurally generated training track segments covering diverse geometric characteristics.}
    \label{fig:training_tracks}
\end{figure}

During training, the policy is exposed to procedurally generated racing tracks with diverse geometric characteristics, as illustrated in Fig.~\ref{fig:training_tracks}. 
The generated layouts include straight-dominant segments, U-turns, single and consecutive hairpins, large-radius curves, and mixed straight--turn transitions. 
Curvature profiles and segment compositions are randomized at each episode to avoid geometry-specific bias.

Training is performed for 10,000 epochs, each consisting of 200 simulation steps. 
Rather than full-lap rollouts, each epoch corresponds to a short-horizon interaction segment of approximately 20~s of simulated time. 
At the beginning of every segment, the arc-length positions of both vehicles are randomly sampled while maintaining a bounded relative distance to ensure active interaction. 
After each segment, the environment is reset with new track instances and initial conditions. 
At every control step, the RL policy outputs high-level interaction parameters, which are passed to the MPC layer to generate dynamically feasible trajectories under friction-circle and actuator constraints.
Experience replay is employed with a buffer size of $10^6$ and a batch size of 256. The key simulation and vehicle parameters are summarized in Table~\ref{table1}.

\begin{table}
\caption{Simulation and Vehicle Parameters}
\label{table1}
\centering
\setlength{\tabcolsep}{3pt}
\renewcommand{\arraystretch}{1.1}
\begin{tabular}{lccc}
\toprule
\textbf{Parameter} & \textbf{Symbol} & \textbf{Value} & \textbf{Unit} \\
\midrule
\multicolumn{4}{l}{\textbf{Simulation Setup}} \\
Time step & $T$ & 0.1 & s \\
Prediction horizon & $N$ & 30 & steps \\
Episode length & -- & 200 steps (20 s) & -- \\
\midrule
\multicolumn{4}{l}{\textbf{Vehicle Model}} \\
Distance from CG to front axle & $l_a$ & 1.38 & m \\
Distance from CG to rear axle & $l_b$ & 1.49 & m \\
Mass & $M$ & 1970 & kg \\
Yaw inertia & $I_z$ & 4760 & kg·m$^2$ \\
Front cornering stiffness & $C_{af}$ & $2.22\times10^5$ & N/rad \\
Rear cornering stiffness & $C_{ar}$ & $2.84\times10^5$ & N/rad \\
Ego vehicle front friction coeff. & $\mu_f^e$ & 1.03 & --\\
Ego vehicle rear friction coeff. & $\mu_r^e$ & 1.08 & --\\
Opponent vehicle front friction coeff. & $\mu_f^o$ & 1.27 & --\\
Opponent vehicle rear friction coeff. & $\mu_r^o$ & 1.34 & --\\
Max velocity & $v_{\max}$ & 35 & m/s \\
\midrule
\multicolumn{4}{l}{\textbf{Constraints}} \\
Steering rate limit & $\dot{\delta}$ & $\pm0.15$ & rad/s \\
Longitudinal jerk limit & $j_x$ & $\pm30$ & m/s$^3$ \\
\midrule
\multicolumn{4}{l}{\textbf{MPC Weights}} \\
Input cost & $R$ & diag(0.01, 0.01) & -- \\
Boundary penalty & $w_{\text{tube}}$ & 10 & -- \\
Collision penalty & $w_{\text{avoid}}$ & 1500 & -- \\
\midrule
\multicolumn{4}{l}{\textbf{RL Settings}} \\
Learning rate (Actor/Critic) & $\alpha,\beta$ & $3\times10^{-4}$ & -- \\
Batch size & -- & 256 & -- \\
Replay buffer size & -- & $10^6$ & -- \\
\bottomrule
\end{tabular}
\end{table}

\begin{figure}[t]
  \centering
  \includegraphics[
      width=0.46\textwidth,
      trim=0 60 0 0,
      clip
  ]{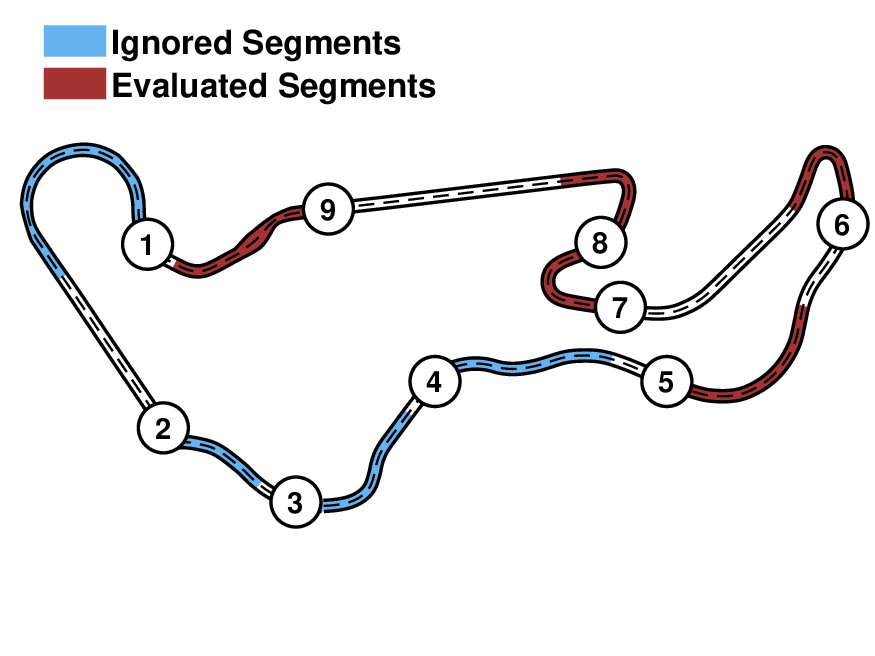}
  \caption{Overall layout and position of the evaluated corners on the Thunderhill West track.}
  \label{fig:track_map}
\end{figure}

The framework is implemented in a custom Python/CasADi closed-loop racing simulator.
For evaluation, the Thunderhill Raceway Park West Circuit (2 miles) is selected as a representative real-world benchmark, as illustrated in Fig.~\ref{fig:track_map}. 
Thunderhill geometry is reserved exclusively for testing and is not used during training. 
Track boundaries are extracted from high-resolution GPS data and processed via spline interpolation to obtain a smooth centerline $C(s)$, from which curvature $\kappa(s)$ is computed. All Frenet quantities are normalized to ensure numerical stability.

In the dynamic layer, both vehicles adopt a nonlinear bicycle model. 
The ego vehicle parameters follow the Lexus LC500 platform reported in \cite{11107568}. 
All actuator limits and friction coefficients reflect physically achievable bounds. 
The MPC layer enforces friction-circle and actuator constraints at each control step, ensuring dynamically feasible behavior independent of the learned policy.

To maintain real-time determinism, the MPC optimization is warm-started from the previous solution and bounded to a 0.1~s computation window. 
If the solver does not improve upon the current solution within this interval, the previously optimized trajectory, which already satisfies the enforced constraints, is retained, preserving constraint-consistent behavior across control cycles. 
In the absence of any prior solution, a neutral control input is applied.

The opponent design is inspired by the interaction reasoning principles in \cite{wang_game-theoretic_2021}, a foundational game-theoretic overtaking framework in autonomous racing, where simplified vehicle models are coupled through iterative equilibrium-based optimization. 
In contrast, our opponent operates directly on the nonlinear single-track vehicle dynamics with friction-circle and actuator constraints enforced at every control step. 
Rather than solving for a coupled equilibrium, the opponent optimizes its own trajectory while incorporating a short preview (1.0 s) of the ego vehicle's predicted motion. 
The objective maximizes longitudinal progress relative to the ego vehicle while penalizing collision risk, boundary violation, and nonsmooth control. 
Combined with elevated friction limits, this results in a dynamically accurate and performance-biased controller that actively initiates overtaking maneuvers whenever feasible.

During simulation, Frenet states, global coordinates, relative distances, control inputs, constraint activations, and boundary or collision events are recorded for replay sampling, convergence analysis, and statistical evaluation 
of defense success rate, collision rate, and trajectory smoothness.

The entire deep RL-MPC framework is implemented in Python on an Alienware Aurora R15 desktop with 13th Gen Intel(R) Core(TM) i9-13900KF and 64 GB RAM. 
The reinforcement learning module is built on \texttt{PyTorch}, while the MPC solver relies on \texttt{CasADi} with an interior-point method.

\begin{figure}[t]
    \centering
    
    \includegraphics[width=0.49\linewidth,trim=0 10 0 40,clip]{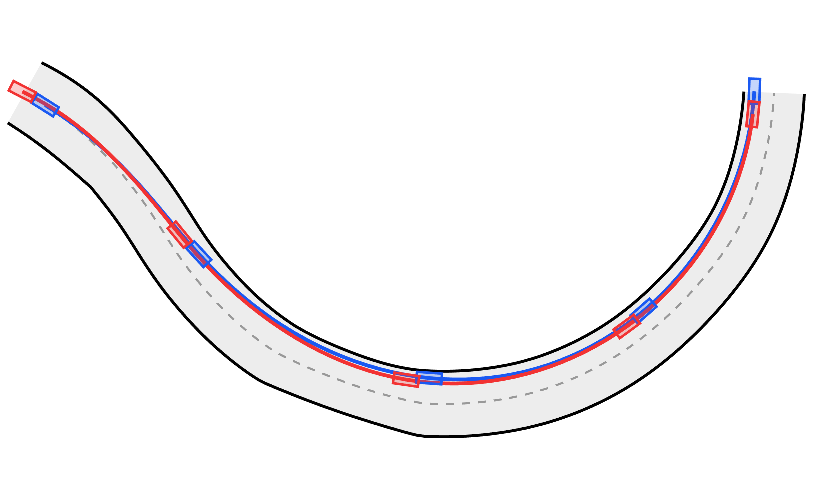}%
    \includegraphics[width=0.49\linewidth,trim=0 10 0 40,clip]{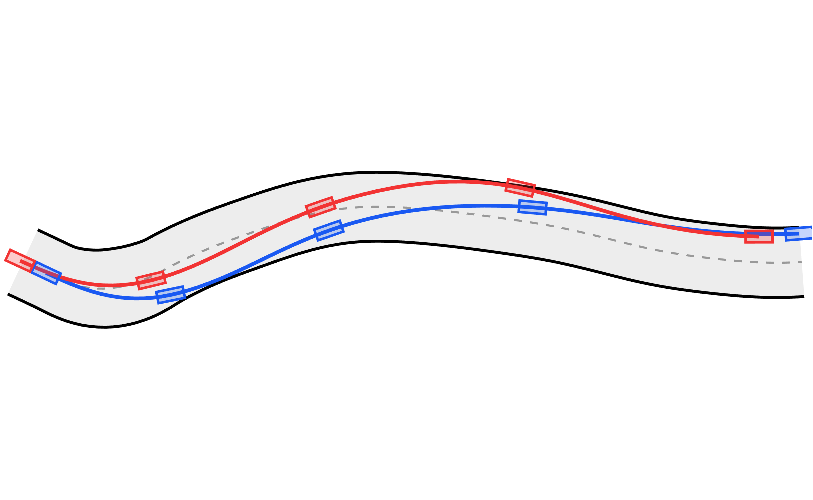}
    
    \vspace{-2.1mm}
    
    \includegraphics[width=0.49\linewidth,trim=0 50 0 40,clip]{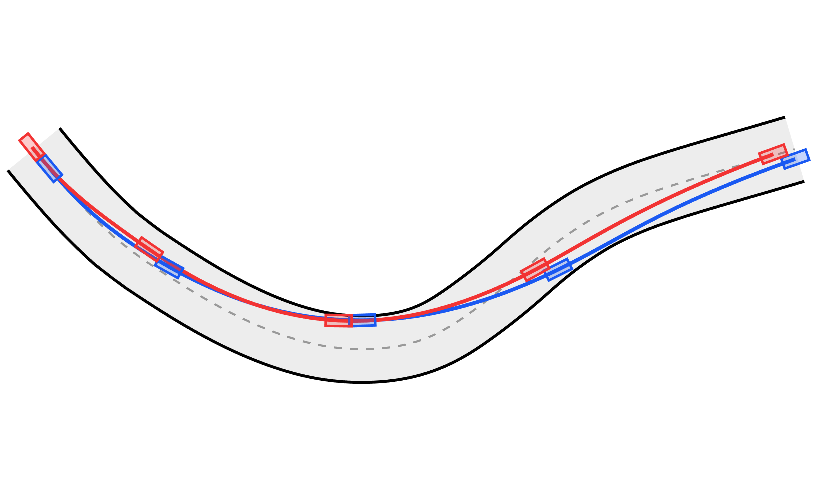}%
    \includegraphics[width=0.49\linewidth,trim=0 50 0 40,clip]{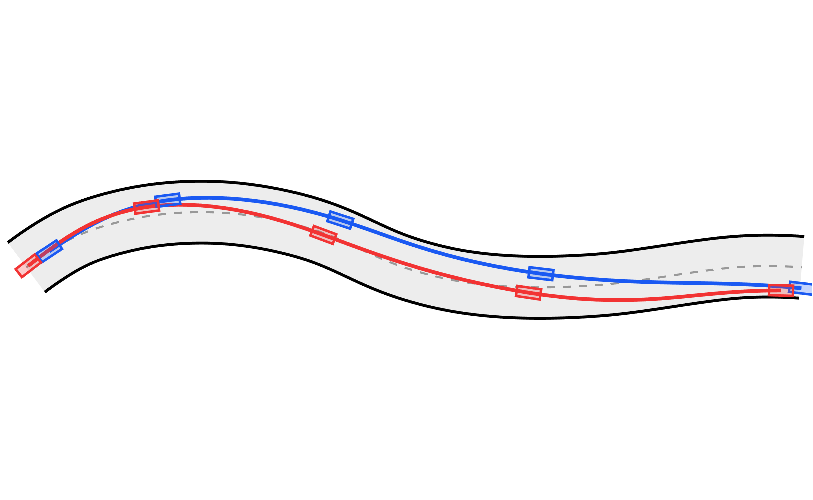}
    
    \caption{Top-down view of ego (blue) and opponent (red) trajectories under the benchmark controller in Segments~1--4. Each segment is rotated for visualization consistency. In all subplots, both vehicles start simultaneously and move from left to right. Apparent vehicle overlap arises from visualization scaling and does not represent physical contact.}
    
    \label{fig:benchmark_seg14}
\end{figure}

\begin{figure*}[t!]
  \centering
  \includegraphics[width=0.96\textwidth]{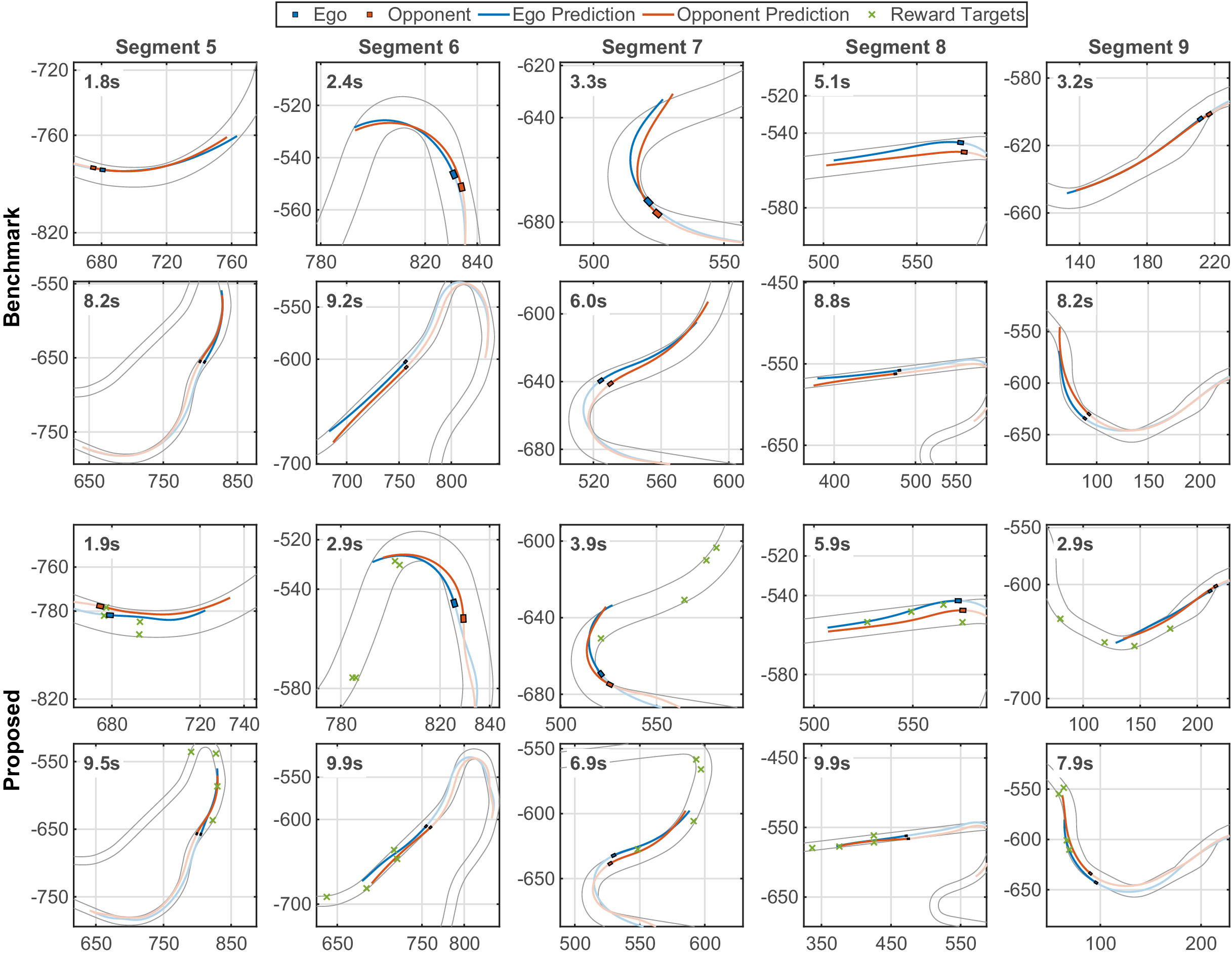}
  \caption{Trajectory-level comparison of ego–opponent interactions between the benchmark and proposed controllers. Rows represent synchronized time frames, aligned such that the opponent’s track progression is identical in both the first and second frames across both trials. Timestamps indicate the elapsed time for each respective frame.}
  \label{fig:trajectory_plots}
\end{figure*}

\subsection{Quantitative Performance Analysis}
\begin{table}[!t]
\centering
\caption{Quantitative Comparison of Defensive Performance at Evaluated Segments}
\label{tab:quantitative_results}

\renewcommand{\arraystretch}{0.95}
\small

\resizebox{\columnwidth}{!}{%
\begin{tabular}{@{}llcccc@{}}
\toprule
Seg. & Ctrl.
& OT (s)$\uparrow$
& Dist (m)$\downarrow$
& Tire (\%)$\uparrow$
& Solve (ms)$\downarrow$ \\
\midrule

\multirow{2}{*}{5}
& Bench. & 14.2 & 549.0 & 88.6 & 23.4 \\
& Prop.  & 15.4 & 498.0 & 86.3 & 31.5 \\

\midrule
\multirow{2}{*}{6}
& Bench. & 7.6 & 503.0 & 88.4 & 27.7 \\
& Prop.  & 20.0 & 483.0 & 86.8 & 34.2 \\

\midrule
\multirow{2}{*}{7}
& Bench. & 6.0 & 501.0 & 78.2 & 22.1 \\
& Prop.  & 13.8 & 455.0 & 83.2 & 34.8 \\

\midrule
\multirow{2}{*}{8}
& Bench. & 7.2 & 585.0 & 62.5 & 23.6 \\
& Prop.  & 17.6 & 539.0 & 70.0 & 36.9 \\

\midrule
\multirow{2}{*}{9}
& Bench. & 8.8 & 550.0 & 88.6 & 25.3 \\
& Prop.  & 6.3 & 549.0 & 90.9 & 29.3 \\

\midrule
\multirow{2}{*}{Avg}
& Bench. & 8.8 & 537.6 & 81.3 & 24.4 \\
& Prop.  & 14.6 & 504.8 & 83.4 & 33.3 \\

\bottomrule
\end{tabular}%
}
\end{table}

The proposed framework is evaluated on the real ThunderHill West track, which is excluded from the training set and used for out-of-distribution testing. The track consists of nine representative segments, as shown in Fig.~\ref{fig:track_map}.

The benchmark controller successfully defends in Segments~1--4. 
In these segments, the baseline MPC solution, combined with local track constraints, happens to restrict feasible passing trajectories. The resulting optimal racing line implicitly occupies the spatial region required for overtaking, thereby preventing the opponent from passing without any interaction-aware mechanism, as shown by the ego and opponent trajectories in Fig.~\ref{fig:benchmark_seg14}.
In contrast, in Segments~5--9, the baseline MPC no longer induces such implicit blocking. The track configuration permits alternative passing corridors, and trajectory optimization alone cannot prevent a faster opponent from overtaking. Therefore, our evaluation focuses on Segments~5--9 to assess the interaction-aware defensive capability of the proposed framework.

Table~\ref{tab:quantitative_results} presents a quantitative comparison between the benchmark controller and the proposed framework across the evaluated segments. 
From left to right, the metrics include the overtake time (i.e., the time before the ego vehicle is overtaken), the opponent’s traveled distance during the 20~s simulation window, the front tire force utilization ratio relative to its theoretical friction limit, and the single-step computational solve time.
Arrows next to the metrics indicate the direction in which the metric corresponds to better performance.

On average, the proposed RL-MPC framework significantly increases the overtake time from 8.8~s to 14.6~s across the evaluated segments, indicating substantially enhanced defensive persistence. 
By actively constraining the opponent’s forward progression, the proposed strategy reduces the opponent’s traveled distance within the fixed 20~s simulation window from 537.6~m to 504.8~m, demonstrating effective suppression of opponent advancement.
Importantly, the front tire utilization, defined as the normalized friction usage derived from the Fiala tire model~\cite{fiala1954seitenkrafte}, remains comparable between the two controllers (83.4\% versus 81.3\%), suggesting that the performance improvement does not stem from artificially exceeding dynamic limits. Instead, the ego vehicle consistently operates near its friction constraints under both strategies.

\begin{figure}[t]
    \centering
    \includegraphics[width=0.9\linewidth]{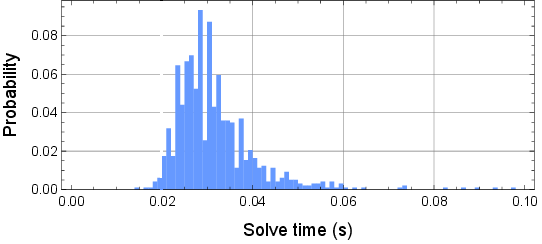}
    \caption{Probability distribution of solve times.}
    \label{fig:solveTimeHistogram}
\end{figure}

While computational requirements increase with complexity, the framework demonstrates real-time feasibility with the implementation of the fallback strategy, achieving mean solve time of 33.3 ms (13.9 ms std) and maximum solve time of 100 ms, within the set 100 ms control interval (Fig.~\ref{fig:solveTimeHistogram}).

\subsection{Qualitative Analysis of Cornering Behaviors}

To evaluate how RL-guided MPC achieves quantitative gains, we analyze simulated trajectories and spatial interactions in Fig. \ref{fig:trajectory_plots}. 

A consistent strategy across trials is the proactive reduction of entry speed; the RL policy places spatial reference points close to the ego vehicle, forcing the MPC layer to slow the vehicle down before the corner.
This compels the opponent to decelerate,  neutralizing its higher tire friction advantage.

In some instances, some spatial reference points are placed side-by-side across the track (e.g., see 1.9~s in Segment 5), because the RL is unaware of the dynamic infeasibility of tracking such reference points.
The MPC layer, however, enforces dynamic feasibility and interprets such references as a strategic decision to reduce speed and move laterally.
It interprets such reference points as a strategic decision to reduce speed and move laterally, thus producing a strategically sound and dynamically safe solution.

\textbf{Defensive Positioning (Segment 5):}
In Segment 5's high-speed sequences, the proposed algorithm maintains a strict inside position and reduces speed ahead of the opponent. 
This forces a corresponding deceleration from the opponent, successfully delaying the opponent's progression by 51~m. 

\textbf{Line Denial and Forcing (Segments 6, 7 \& 8):} In Segment 6 and 7, the benchmark takes an early apex and is overtaken. 
Conversely, the proposed method holds the inside and pushes the opponent toward the track exterior, forcing a high-radius, inefficient path for the opponent. 
In Segment 6 and 8, the ego vehicle maintains its lead through the hairpin and defends the subsequent straight by forcing the opponent to cover greater lateral distances using lateral movements.

\textbf{Segment 9 Performance:} Segment 9 represents an exception to the overall trend. A possible explanation is that this segment creates a particularly difficult defensive situation, where the opponent still has a viable passing corridor even when the ego vehicle adjusts its reference. In this case, the learned policy may not identify an effective blocking strategy, so the generated references alter the ego trajectory without substantially limiting the opponent’s overtaking opportunity.

\section{Conclusion}
\label{sec:conclusions}

This paper introduces a hierarchical racing framework that integrates a high-level RL strategic planner with a low-level nonlinear MPC for defensive blocking. 
In simulated trials, the agent successfully delays its opponent by a total of 5.8~s compared to a non-interactive baseline, demonstrating strategic behaviors such as preemptive apex deceleration and defensive positioning against aggressive late-braking overtaking attempts.
Furthermore, the framework utilizes over 83\% of available tire force (benchmark 81.3\%) and achieves a 33.3 ms average solve time (13.9~ms std).

Several limitations remain. 
The current evaluation is simulation-only, uses one held-out real track, and compares against Spatial Envelope MPC without learned references as the benchmark. 
Future work should validate the controller on hardware, test multiple tracks and opponent policies, add measurement noise and latency, compare against additional interaction-aware defensive baselines, and encode racing-rule constraints on unsafe or excessive blocking.

\bibliographystyle{IEEEtran}
\bibliography{refs}

@article{huang_fair_2025,
  author  = {Z. Huang and C. Hao and W. Zhan and J. Ma and M. Tomizuka},
  title   = {Fair Play in the Fast Lane: Integrating Sportsmanship into Autonomous Racing Systems},
  journal = {arXiv:2503.03774},
  year    = {2025}
}

@InProceedings{drews_aggressive_2017,
  title = 	 {Aggressive Deep Driving: Combining Convolutional Neural Networks and Model Predictive Control},
  author = 	 {Drews, Paul and Williams, Grady and Goldfain, Brian and Theodorou, Evangelos A. and Rehg, James M.},
  booktitle = 	 {Conference on Robot Learning},
  pages = 	 {133--142},
  year = 	 {2017},
  volume = 	 {78},
  skip_series = 	 {Proceedings of Machine Learning Research},
  skip_publisher =    {PMLR},
}

@article{rickenbach_zipmpc_2025,
  author  = {R. Rickenbach and A. A. Lahoud and E. Schaffernicht and M. N. Zeilinger and J. A. Stork},
  title   = {{ZipMPC}: Compressed Context-Dependent {MPC} Cost via Imitation Learning},
  journal = {arXiv:2507.13088},
  year    = {2025}
}

@article{gros_data-driven_2020,
  author  = {S. Gros and M. Zanon},
  title   = {Data-Driven Economic {NMPC} Using Reinforcement Learning},
  journal = {IEEE Trans. Autom. Control},
  volume  = {65},
  number  = {2},
  pages   = {636--648},
  year    = {2020},
  doi     = {10.1109/TAC.2019.2913768}
}

@article{amos_differentiable_2019,
  author  = {B. Amos and I. Jimenez and J. Sacks and B. Boots and J. Z. Kolter},
  title   = {Differentiable {MPC} for End-to-End Planning and Control},
  journal = {Adv. Neural Inf. Process. Syst.},
  volume  = {31},
  year    = {2018}
}

@article{costa_online_2023,
  author  = {G. Costa and J. Pinho and M. A. Botto and P. U. Lima},
  title   = {Online Learning of {MPC} for Autonomous Racing},
  journal = {Robot. Auton. Syst.},
  volume  = {167},
  pages   = {104469},
  year    = {2023}
}

@article{wang_game-theoretic_2021,
  author  = {M. Wang and Z. Wang and J. Talbot and J. C. Gerdes and M. Schwager},
  title   = {Game-Theoretic Planning for Self-Driving Cars in Multivehicle Competitive Scenarios},
  journal = {IEEE Trans. Robot.},
  volume  = {37},
  number  = {4},
  pages   = {1313--1325},
  year    = {2021}
}

@article{subosits_racetrack_2019,
  author  = {J. K. Subosits and J. C. Gerdes},
  title   = {From the Racetrack to the Road: Real-Time Trajectory Replanning for Autonomous Driving},
  journal = {IEEE Trans. Intell. Veh.},
  volume  = {4},
  number  = {2},
  pages   = {309--320},
  year    = {2019}
}

@inproceedings{haarnoja_soft_2018,
  author    = {T. Haarnoja and A. Zhou and P. Abbeel and S. Levine},
  title     = {Soft Actor-Critic: Off-Policy Maximum Entropy Deep Reinforcement Learning with a Stochastic Actor},
  booktitle = {International Conference on Machine Learning},
  pages     = {1861--1870},
  year      = {2018},
  skip_organization = {PMLR}
}

@article{yu2024real,
  title={A Real-Time Terrain-Adaptive Local Trajectory Planner for High-Speed Autonomous Off-Road Navigation on Deformable Terrains},
  author={Yu, Siyuan and Shen, Congkai and Dallas, James and Epureanu, Bogdan I and Jayakumar, Paramsothy and Ersal, Tulga},
  journal={IEEE Trans. Intell. Transp. Syst.},
  year={2024},
  publisher={IEEE}
}

@article{yu_spatial_2025,
	title = {Spatial {Envelope} {MPC}: {High} {Performance} {Driving} without a {Reference}},
	publisher = {arXiv},
	author = {Yu, Siyuan and others},
	journal = {arXiv:2509.18506},
	year = {2025},
}

@article{tubempcracing,
  author={Wischnewski, Alexander and Herrmann, Thomas and Werner, Frederik and Lohmann, Boris},
  journal={IEEE Trans. Intell. Veh.}, 
  title={A Tube-{MPC} Approach to Autonomous Multi-Vehicle Racing on High-Speed Ovals}, 
  year={2023},
  volume={8},
  number={1},
  pages={368-378},
}

@article{YU2021412,
title = {Nonlinear Model Predictive Planning and Control for High-Speed Autonomous Vehicles on {3D} Terrains},
journal = {IFAC-PapersOnLine},
volume = {54},
number = {20},
pages = {412-417},
year = {2021},
skip_note = {Modeling, Estimation and Control Conference},
issn = {2405-8963},
author = {Siyuan Yu and Congkai Shen and Tulga Ersal},
}

@article{neuralnetMPCFriction,
  author={Spielberg, Nathan A. and Brown, Matthew and Gerdes, J. Christian},
  journal={IEEE Trans. Control Syst. Technol.}, 
  title={Neural Network Model Predictive Motion Control Applied to Automated Driving With Unknown Friction}, 
  year={2022},
  volume={30},
  number={5},
  pages={1934-1945}}

@article{wurman_outracing_2022,
  author  = {Wurman, Peter R. and others},
  title   = {Outracing champion {Gran Turismo} drivers with deep reinforcement learning},
  journal = {Nature},
  volume  = {602},
  number  = {7896},
  pages   = {223--228},
  year    = {2022}
}

@article{hu2024game,
  title={A Game Theoretic Method for Two-Team Multi-Player Autonomous Racing},
  author={Hu, Zhenghao and Li, Xiuxian and Meng, Min and Zhao, Shiyu},
  journal={IEEE Robot. Autom. Lett.},
  volume={9},
  number={9},
  pages={7581--7588},
  year={2024},
  publisher={IEEE}
}

@INPROCEEDINGS{11127283,
  author={Hu, Haimin and Fisac, Jaime Fernández and Leonard, Naomi Ehrich and Gopinath, Deepak and DeCastro, Jonathan and Rosman, Guy},
  booktitle={International Conference on Robotics and Automation}, 
  title={Think Deep and Fast: Learning Neural Nonlinear Opinion Dynamics from Inverse Dynamic Games for Split-Second Interactions}, 
  year={2025},
  volume={},
  number={},
  pages={16678-16684},
  doi={10.1109/ICRA55743.2025.11127283}}

@article{lidard2024blending,
  title={Blending data-driven priors in dynamic games},
  author={Lidard, Justin and others},
  journal={arXiv:2402.14174},
  year={2024}
}

@article{remonda2024simulation,
  title={A simulation benchmark for autonomous racing with large-scale human data},
  author={Remonda, Adrian and others},
  journal={Adv. Neural Inf. Process. Syst.},
  volume={37},
  pages={102078--102100},
  year={2024}
}

@INPROCEEDINGS{11107568,
  author={Suminaka, Makoto and Dallas, James and Thompson, Michael and Soga, Masayuki and Kasai, Eiji and Subosits, John},
  booktitle={American Control Conference}, 
  title={Adaptable High-Speed Model Predictive Control for Autonomous Drifting: Koopman-Based Dynamics}, 
  year={2025},
  volume={},
  number={},
  pages={1057-1063},
  doi={10.23919/ACC63710.2025.11107568}}

@article{fiala1954seitenkrafte,
  title={Seitenkräfte am rollenden {L}uftreifen},
  author={Fiala, E},
  journal={VDI Zeitschrift},
  volume={96},
  pages={973-979},
  year={1954}
}

@inproceedings{song2021overtaking,
  author    = {Y. Song and H. Lin and E. Kaufmann and P. D{\"u}rr and D. Scaramuzza},
  title     = {Autonomous Overtaking in {Gran Turismo Sport} Using Curriculum Reinforcement Learning},
  booktitle = {International Conference on Robotics and Automation},
  pages     = {9403--9409},
  year      = {2021},
  address   = {Xi'an, China},
  doi       = {10.1109/ICRA48506.2021.9561049}
}

@article{hori2025learning,
  author  = {T. Hori and J. DeCastro and D. Gopinath and A. Balachandran and G. Rosman},
  title   = {Learning to Plan, Planning to Learn: Adaptive Hierarchical {RL-MPC} for Sample-Efficient Decision Making},
  journal = {arXiv:2512.17091},
  year    = {2025},
  doi     = {10.48550/arXiv.2512.17091}
}

\end{document}